# Deep Learning for Multi-Label Learning

## A Comprehensive Survey

Adane Nega Tarekegn, Mohib Ullah, *Member, IEEE,* Faouzi Alaya Cheikh, *Senior Member, IEEE*

*Abstract*—Multi-label learning is an essential component of supervised learning that aims to predict a list of relevant labels for a given data point. In the era of big data, characterized by the continuous generation of complex datasets, multi-label learning tasks, such as multi-label classification (MLC) and multi-label ranking, present significant challenges, capturing considerable attention across various domains. Some of the inherent challenges include high-dimensional features and labels, label dependency, and the existence of partial or missing labels, all of which render traditional methods ineffective. In recent times, there has been a notable surge in the adoption of deep learning (DL) techniques to address these challenges more adeptly in MLC. Notably, there is a burgeoning effort to harness the robust learning capabilities of DL for improved modelling of label dependencies and other inherent complexities in MLC. However, it is noteworthy that comprehensive studies exclusively dedicated to DL for multi-label learning remain scarce. Hence, this survey aims to meticulously review recent advancements in DL for multi-label learning while also offering a concise overview of open research issues in MLC. The survey consolidates existing research efforts in DL for MLC, such as deep neural networks, transformers, autoencoders, and convolutional and recurrent architectures. Finally, the study provides a comparative analysis of existing approaches to furnish insightful observations and provoke future research trajectories in this domain.

## I. INTRODUCTION

IN many real-world applications, an object may be associated with multiple labels concurrently, and such problems are recognized as multi-label learning (MLL) [1]. MLL extends the conventional single-label learning approach, where there is typically a finite set of potential labels that can be applied to the instances of multi-label data (MLD). The basic goal is to simultaneously predict a vector of outputs for a given single input, which means that it is possible to solve more complex decision-making problems. This is opposed to the standard single label classification, in which each instance is assigned only one label.

In the context of multi-label tasks, an instance is typically associated with a set of labels, constituting distinct combinations known as relevant labels (active labels), while labels not linked to the instance are termed irrelevant labels. The quantity of labels within the MLD determines the size of the binary vector representing both relevant and irrelevant labels. Depending on the goal, there are two primary tasks in MLL: multi-label classification (MLC) and multi-label ranking (MLR) [2]. MLC constitutes the primary learning task, aiming to train a model that segregates the label set into relevant and irrelevant categories with relative to a query instance. On the other hand, MLR focuses on training a model to arrange the class labels based on their relevance to a query instance.

Although MLC applications traditionally concentrate on text analysis, multimedia, and biology, their significance is progressively growing across diverse domains, such as document classification [3][4][5], healthcare [6][7][8], environmental modelling [9][10], emotion recognition [11][12], commerce [13][14], social media [15][16][17], and more. Many other demanding tasks, including annotation of videos, language modelling, and categorizing web pages can also derive benefits by framing them as MLC tasks involving hundreds or thousands of labels. Such extensive label space pose study challenges, including issues related to data sparsity and scalability. MLC has additional complexities, including modelling label correlations [18][19], imbalanced labels [20] and nosily labels [21]. Traditional MLC approaches, such as algorithm adaptation and problem transformation [22][23], demonstrate suboptimal performance in addressing these challenges.

Apart from the traditional approaches, deep learning (DL) techniques have gained increasing popularity in addressing the challenges of MLC. The formidable learning capacities of deep learning are particularly adaptable for addressing MLC challenges, as demonstrated by their notable success in addressing single-label classification tasks. Currently, a predominant trend in MLC involves extensively incorporating DL techniques even for more challenging problems, such as Extreme MLC [24][25][26], imbalanced MLC [27][28], weakly supervised MLC [29][30][31], and MLC with missing labels [32][33]. Effectively harnessing the strong learning capabilities of DL is crucial to better understand and model the label correlations, thereby enabling DL to tackle MLC problems effectively. Several studies have shown that MLC methods explicitly designed to capture label dependencies typically demonstrate superior predictive performance [34][19]. This paper conducted a concise review of the existing literature to identify a broad range of DL-based techniques for MLC problems to inspire further exploration of innovative DL-based approaches for MLC. Few surveys are already available on traditional approaches for MLC, such as those referenced in [35][23] [36]. Additionally, there are surveys that contain both traditional and DL methods [37][38], but these have limited coverage of the state-of-the-art DL approaches for MLC and are focused on specific domains. However, this paper uniquely concentrates on a range of DL architectures, including recurrent and convolutional networks, transformers, autoencoders, and hybrid models, for addressing MLC challenges across diverse domains. Fig. 1 presents a taxonomy of MLL methods comprising both traditional approaches and DL methods.

The main contributions can be outlined as follows:
1. This survey thoroughly covers DL methods for solving MLC tasks, involving different domains and data modalities, including texts, music, images, and videos.
2. A comprehensive summary of the most recent DL methods for MLC across several publicly available datasets is provided (Tables I, II, and III), with a brief overview of each DL method and insightful discussions. Therefore, readers can better identify the constraints of each approach, thereby facilitating the development of more advanced DL methods for MLC.
3. We have provided a brief description of the current challenges facing the realm of MLC. Additionally, we have included a summary of the multi-label datasets utilized in MLC, along with the definition of attributes used to evaluate the characteristics of these datasets.
4. Finally, this paper provides a comparative study of existing approaches involving various DL techniques and examines the advantages and disadvantages of each method (Table V). It offers insights that can guide the selection of suitable techniques and develop better DL approaches for MLC in future studies.

The subsequent sections of the paper are organized as follows. Section II presents the fundamental concepts of multi-label learning. Section III presents the research methodology, which focuses on the data source and search strategy, selection criteria, and statistical trends of publications. Section IV serves as the focal point of this survey, exploring a range of DL approaches for tackling the MLC challenges. Section V focuses on open challenges in MLC and datasets. Section VI offers the comparison of various methods, outlining their respective advantages and limitations. Finally, section VII provides the conclusions of the paper.

## II. FUNDAMENTAL CONCEPTS OF MLL

MLL is founded on a dataset where instances are associated with several target variables or labels simultaneously. The main goal when working with such data is MLC, which aims to categorize the target variables into relevant and irrelevant groups for a specific instance. Additional tasks may include ranking the labels according to their relevance or creating a comprehensive joint distribution encompassing all possible assignments of values to the labels. The formal definition of MLL can be presented as follows [39]: Let $X$ be a $d$-dimensional input space of categorical or numerical features and output space of $q$ labels $L = \{\lambda_1, \lambda_2, ..., \lambda_q\}, q > 1$. A multi-label example can be defined as a pair $(x, Y)$ where $x = (x_1, x_2, ..., x_q) \in X$ and $Y \subseteq L$ is called a label-set. $D = \{(x_i, Y_i) \mid 1 \leq i \leq m\}$ is an MLD composed of a set of $m$ instances. Let Q be a quality criterion that rewards models with high predictive performance and low complexity. If the task is an MLC, then the aim is to compute a function $h: X \rightarrow 2^L$ such that $h$ maximizes $Q$. If the task is an MLR, then the goal is to find a function $f: XxL \rightarrow \mathbb{R}$ such that $f$ maximizes $Q$, where $\mathbb{R}$ is the ranking of labels for a given sample.

MLC is currently receiving significant attention and is applicable to a variety of research domains, including bioinformatics [40][41], text classification [42][43], music categorization [44][45], medical diagnosis [46][47], image classification [48], and video annotation [49]. For example, in medical diagnosis, it's common for a patient to experience multiple side effects associated with a disease, or a medical diagnosis might find a patient suffering from more than one disease at the same time. An image from the real world can be assigned multiple labels due to its inherent richness in semantic information, encompassing objects, scenes, actions, attributes, and their interactions. Effectively modeling this diverse semantic information and its interdependencies is crucial for comprehensive image understanding. In-text categorization, a news article may include several facets of an event, leading to its assignment under a diverse array of topics. In such scenarios, the objective is to allocate a label set to every new instance encountered [50]. In MLR, the goal is to not only predict a vector of outputs from a finite set of predefined labels but also to rank them based on their relevance to the provided input. In a multi-label learning scenario, the task extends beyond predicting relevant and irrelevant labels; it often involves generating a well-ordered ranking of relevant labels (i.e., a list of preferences) from the list of possible labels for each unseen example. MLR is an interesting problem as it subsumes many supervised learning tasks such as multi-label, multi-class, and hierarchical classifications [51]. Document classification is a prominent use case for MLR, involving the categorization of topics (such as technology, politics, and sports) within a collection of documents, such as news articles. It's common for a document to be associated with multiple topics, and the aim of the learning algorithm is to rank (order) the relevant topics higher than non-relevant ones for a given document query.

Two traditional approaches exist for solving the MLL task: algorithm adaptation and problem transformation. Algorithm adaptation aims to modify or extend conventional learning methods learning to directly handle MLD [53]. On the other hand, problem transformation involves converting the MLC task into either one or multiple single-label classification tasks [52], or label ranking tasks [2]. The three most prominent methods from the problem transformation category include label power-set (LP) [54], binary relevance (BR) [1], and classifier chains (CC) [55]. The BR method breaks down the multi-label problem into a series of independent binary problems. Subsequently, each binary problem is addressed using a traditional classifier. Finally, the individual predictions are combined to get the subset of labels relevant to each test instance. Although BR is relatively simple to implement, it is realized that BR ignores the possible relationship between labels (such as label dependency, cooccurrence, and correlation). To deal with the limitation of the BR method, the classifier chain (CC) [55] was introduced. This method interconnects binary classifiers in a sequential chain, where the predictions of preceding classifiers serve as features for subsequent classifiers. This allows the latter classifiers to leverage the correlation with earlier predictions to enhance the quality of their predictions. The approach of LP involves treating every distinct label combination as a class identifier, thereby converting the original MLD into a multi-class dataset. Following the training of a conventional classifier with this dataset, the predicted classes are then converted back into the respective label subsets. Both LP and CC are traditional approaches to learning the interdependency between labels for MLL. However, in addition to high computation cost when dealing with a larger number of labels, CC and LP have limited ability to capture the high-order correlations between labels.

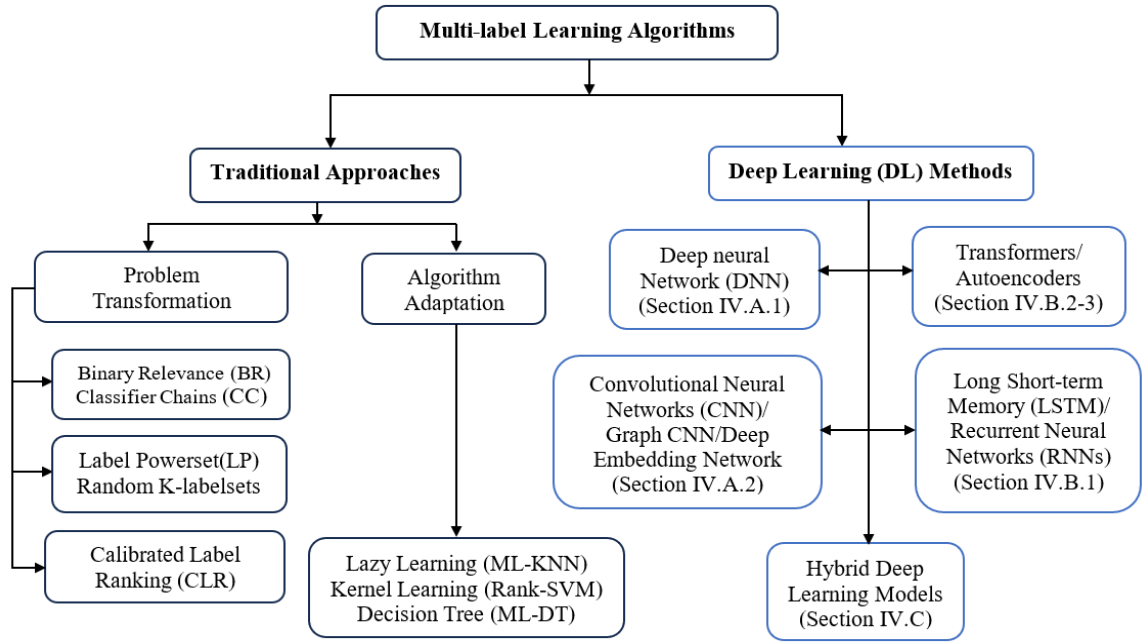

Fig. 1. A taxonomy of multi-label learning methods encompassing both traditional and deep learning methods.

Regardless of the methods to address the multi-label problem and the approaches to the label correlations, MLC has additional complexities, such as grappling with higher-order dependencies among labels, contending with an extensive array of labels requiring substantial computational resources, and handling scenarios involving partially or weakly supervised MLC as well as imbalanced MLC [20]. Moreover, the classical approaches mentioned earlier prove ineffective in addressing these challenges. Recently, deep learning (DL) techniques have gained increased popularity across diverse disciplines, and MLC has been no exception to benefiting from the latest developments in DL. Therefore, the objective of this survey is to provide a thorough review of the DL methods for MLC that aims to address these challenges and promote the application of DL-based MLC in various domains.

## III. METHODOLOGY

This section reveals the search strategy, study selection criteria, and trends in publication to ensure a thorough and objective selection of literary sources.

### A. Search Strategy

In this comprehensive survey, an exploration of research papers pertaining to DL approaches for MLC, including publications from 2006 to 2023, was conducted. Initially, prominent library databases spanning various research domains were used as primary sources: Springer, IEEExplore, DBLP, ACM Digital Library, Science Direct, and Google Scholar, among others. Boolean operators were employed to refine searches, amalgamating terms with synonymous meanings and constraining the scope of inquiry. Predetermined search terms, incorporating phrases, such as 'deep learning for multi-label classification,' 'multi-label classification using deep convolutional neural networks (CNN),' 'multi-label prediction using recurrent neural network (RNN), transformers, autoencoders', or 'hybrid deep learning for multi-label classification,' were applied. Additionally, efforts were made to identify relevant articles from alternative sources, including peer-reviewed journals and conferences.

### B. Selection Criteria

This paper primarily focuses on examining DL techniques for addressing MLC. We established a set of eligibility criteria, all of which needed to be met simultaneously, for selecting relevant publications: (1) the publication on MLL should contain MLD and DL; (2) it either employs or suggests DL methods for solving MLC; (3) experimental results evaluate DL methods using metrics for multi-label scenarios; (4) a full-text article written in English. Publications proposing DL methodologies for addressing MLC are included for review without any restrictions on publication dates. A total of 382 publications were initially gathered and identified during the search process. Out of these, 64 duplicates were identified, and 106 were excluded after screening titles and abstracts. Subsequently, a thorough examination of the full text of each paper was conducted, resulting in the identification of 212 relevant papers for inclusion in this study. Additionally, any publications with duplicated titles, abstracts, or content were carefully removed, ensuring the retention of only one copy of each publication in the final selection.

*C. Publication Trends*

Approximately 15 years ago, the landscape of MLC started to attract researchers, marking the emergence of this dynamic field as a compelling research topic. Fig. 2 shows an increasing trend of publications related to DL-based MLC, covering the period from 2006 to 2023. Notably, the number of publications demonstrated a consistent increase from 2012 to 2023. In 2019, there was a slight dip in publicati
ons compared to 2018; however, subsequent years displayed an upward trajectory. Particularly in recent times, the volume of published works on MLC employing DL techniques has significantly surpassed that of previous years. This observation underscores the continued exploration of innovative DL techniques to tackle MLC tasks as a noteworthy and actively researched area, attracting substantial attention and interest from the research community.

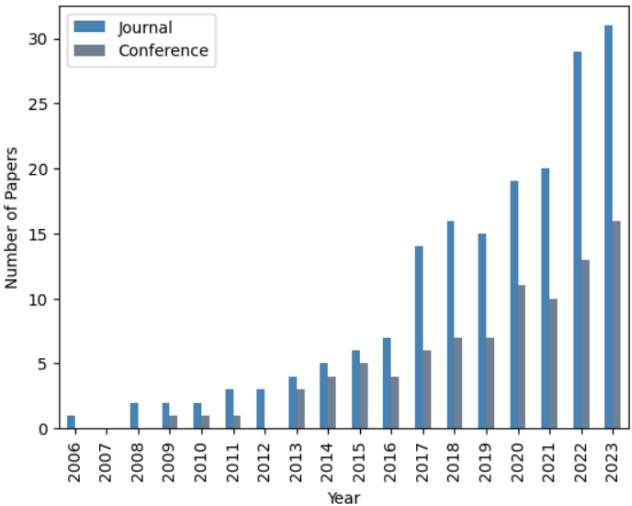

Fig. 2. The quantity of publications related to the topic of MLC utilizing DL methods according to Google Scholar sources from 2006-2023.

## IV. DEEP LEARNING FOR MLC

Recent advances in DL have significantly enriched the landscape of MLC. DL architectures play a pivotal role in generating embedding representations for both input features and output space. The formidable learning capabilities of DL find widespread application in MLC tasks in various domains, such as images, textual, music, and videos. The most frequently used DL methods for MLC include deep neural networks, convolutional, recurrent, autoencoder and transformer architectures, and hybrid models. Effectively harnessing the advantages of these DL approaches is crucial in addressing label dependencies and other challenges in MLC. This section provides an overview of these prominent DL methods for MLC with an overview and detailed examination of each technique specifically tailored for MLC.

*A. Neural Networks for MLC*

This section provides an in-depth exploration of deep neural networks (DNN) and convolutional neural networks (CNN) for MLC, along with a summary of the most recent DL methods, applications, and datasets in MLC.

**1) Deep Neural Networks for MLC**

Deep neural networks (DNNs) have been employed to address MLC problems, and the simplest approach is to decompose the MLC problem into several sets of binary classification problems, one for each label. However, this strategy encounters scalability issues, particularly when handling a substantial number of labels. Additionally, it considers missing labels as negatives, resulting in a performance decline, and ignores dependencies among labels, which is an important aspect of effective recognition. Therefore, a different approach that focuses on the use of label relationships needs to be explored. One such approach is BP-MLL (Backpropagation for Multi-label Learning) [56], which frames MLC problems as a neural network featuring numerous output nodes, with each node representing a distinct label. BP-MLL is applied to an MLC employing logistic activation neurons with a single hidden layer, supplemented by biases originating from both the input and hidden layers. The input layer size corresponds to the total number of available features (plus a bias neuron). Let $X = \mathbb{R}^d$ denote a $d$-dimensional input space and $L = \{\lambda_1, \lambda_2, …, \lambda_q\}, q > 1$ be an output space of $q$ labels, $D = \{(x_i, Y_i) \mid 1 \leq i \leq m\}$ is an MLD comprising a set of $m$ instances, where each instance $x_i \in X$ is depicted as a d-dimensional feature vector, along with a set of q labels linked to the feature vector and $Y_i \subseteq L$ is a label-set. A Multi-Layer Perceptron (MLP) neural network, as depicted in Fig. 3, can be employed, which includes d input neurons corresponding to the d-dimensional feature vector and q output neurons representing the output labels. The size of the output layer matches the number of labels. The network architecture involves d×n weights ($W_{ih}, 1 \leq i \leq d, 1 \leq h \leq n$) connecting the input and hidden layers, and n×q weights ($W_{ho}, 1 \leq h \leq n, 1 \leq o \leq q$) connecting the hidden and output layers. Additionally, bias parameters are denoted as $I$ for the input layer and $H$ for the hidden layer. Given that the objective of MLL is to predict labels for test samples, it is imperative to assess the global error of the model (Eq. 1).

$$E = \sum_{i=1}^{m} E_i \quad (1)$$

The error, $E_i$, pertains to the sample $x_i$ and can be articulated as:

$$E = \sum_{j=1}^{Q} \left(c_j^i - d_j^i\right)^2 \quad (2)$$

Here, $c_j^i = c_j(x_i)$ represents the $j^{th}$ predicted label on the instance $x_i$, while $d_j^i$ denoting the $j^{th}$ true label of the instance $x_i$. The true label has a value of either $+1 (j \in Y)$ or $-1 (j \notin Y)$.

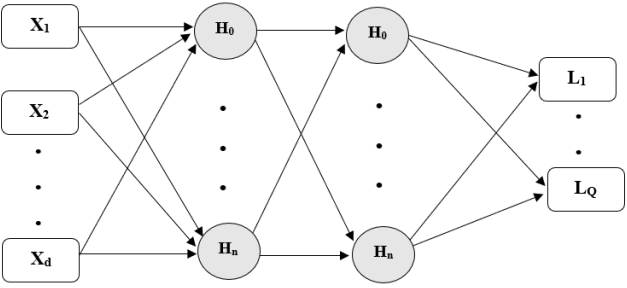

Fig. 3. Structure of the neural network for MLC

Various learning algorithms are applicable for acquiring a model from a training dataset. The algorithm of backpropagation, for instance, employs the process of error-based learning. Nevertheless, it may not be suitable for MLL due to the neglect of label correlations in the error function (2). In the original backpropagation algorithm, the error function (2) restricts the discrimination of individual labels, whether a particular label $j \in L$ is part of the sample $x_i$ or not.

One should consider the labels within $Y_i$ hold greater significance compared to those outside of $Y_i$. BP-MLL views each output node as a binary classification problem, and training is based on the classical BP algorithm, but in order to address the dependencies across labels, the new global error function is proposed that relies on the output layer to minimize pairwise raking errors.

$$E = \sum_{i=1}^{m} E_i = \sum_{i=1}^{m} \frac{1}{|Y_i||\widehat{Y_i}|} \sum_{(k,l) \in Y_i \times \widehat{Y_i}} \exp\left(-\left(C_k^i - C_l^i\right)\right) \quad (3)$$

In equation (3), the error function accumulates the errors for each sample $(x_i, Y_i)$, where $\widehat{Y_i}$ represents the complementary set of $Y_i$ in L, and |.| calculates the cardinality of a set. The term $c_k^i - c_l^i$ denotes the disparity between predicted and actual labels. This error term sums the differences between label pairs: those belonging to the sample and those that do not. Normalization is done by the total number of pairs, $|Y_i||\widehat{Y_i}|$. Subsequently, correlations between label pairs are computed. The error function quantifies label output disparities. Learning entails minimizing this function by amplifying output values for labels belonging to training samples and reducing those for non-members. When training data adequately covers the sample space distribution, the model learns by minimizing the error function through training sample input. The subsequent stage in attaining the MLC classifier involves identifying the label set associated with the input instance, which can be extracted from the output values of the neural network using the threshold function. If the value of the output neuron surpasses the threshold, the respective label is attributed to the input instance; otherwise, it is not.

BP-MLL pioneered the use of neural network architectures for tackling MLC, leveraging their capability to incorporate label correlations. This approach is expected to outperform traditional neural networks in multi-label scenarios where considering label dependencies is crucial. However, it is found that BM-LL requires more computational complexity and convergence speed as the number of labels grows. As a result, it was later extended by state-of-the-art learning techniques. The authors in [57] proposed an improvement to the BP-MLL method by modifying the global error function. This modified error function allows the threshold value to be determined automatically by adaptation during neural network learning instead of using an additional step in BP-MLL to define the threshold function. Furthermore, [58] found the suboptimal performance of BP-MLL on textual datasets. In response to this limitation, [58] explored the constraints of BP-MLL by substituting ranking loss minimization with the more commonly employed cross-entropy error function. The authors demonstrate the capability of a single hidden layer neural network to reach cutting-edge performance levels in extensive multi-label text classification assignments by leveraging the available techniques in DL, such as ReLUs, AdaGrad, and Dropout.

In a different study documented in [59], a label-decision module was integrated into DNNs, resulting in the attainment of top-tier accuracy in multi-label image classification tasks. Building upon this framework, Du et al. [60] introduced ML-Net, a DNN designed for the MLC of biomedical texts. ML-Net incorporates the label–decision module from [59], but it converts the framework from image processing to text classification. The ML-Net model integrates label prediction and decision-making within the same network, enabling the determination of output labels through a combination of label confidence scores and document context. Its objective is to reduce pairwise ranking errors among labels, allowing for end-to-end training and prediction of the label set without requiring an additional step for determining output labels.

Recently, [61] proposed a new loss for MLC, named ZLPR loss, to extend the application of DL in MLC. The authors extended the cross-entropy loss from the single-label classification, which is expressed in Eq. (4).

$$Loss_{zlpr} = \log\left(1 + \sum_{i \in \Omega_{pos}} e^{-s_i}\right) + \log\left(1 + \sum_{i \in \Omega_{neg}} e^{s_j}\right) \quad (4)$$

where $\Omega_{poss}$ denotes the set of outputs, $\Omega_{neg} = \Lambda/\Omega_{pos}$, $s_i$ representing the score of the MLC model for the $i\_th$ the category ($\lambda_i$). In contrast to earlier ranking-based losses, ZLPR exhibits the capability to dynamically determine the number of target categories while enhancing a model's label-ranking proficiency. In comparison to certain binary losses, the ZLPR loss excels in capturing a more robust correlation of labels and elucidating the ranking relationship between negative and positive categories.

**Remarks**: DNNs have been one of the most widely used DL methods for MLC. To support the application of DNN for MLC tasks, various loss functions have been developed to determine the range of tasks. BP-MLL loss, identified as one of the early studies, is acknowledged as one of the first DNNs for MLC, undergoing subsequent enhancements by a variety of researchers. The application of MLC using DNN architecture has been applied in various areas. In the healthcare domains, for instance, it has been used for tasks such as intelligent health risk prediction [62], protein function prediction [63], encoding electronic medical records [64], and multi-label chronic disease prediction [65]. Other related tasks using DNN for MLC include SLA violation prediction [66], hierarchical DNN for peptide bioactivities [67], and Robust DNN for multi-label image classification [68].

2) **Deep CNN for MLC**

Deep CNNs have shown encouraging results in a single-label image learning problem. However, multi-label image classification represents a broader and more practical challenge, as most real-world images contain objects belonging to multiple distinct categories. The success demonstrated by deep CNN-based models in single-label image classification can be expanded and applied to tackle multi-label challenges. The multi-label task is addressed by improving the DL models from an architectural viewpoint, in particular, the loss layer. To establish multi-label losses, the research endeavour is primarily devoted to improving binary cross-entropy (BCE). A study by [69] delved into various multi-label losses, including SoftMax, Pairwise Ranking, and weighted approximated-ranking loss (WARP), when training CNN. The findings indicated that the WARP loss (Eq. (5)) works well in addressing multi-label annotation challenges.

$$L_{warp} = \sum_{u \notin Y_i} \sum_{u \in Y_i} w(r_i^u) \max(0, \alpha + f_u(x_i) - f_u(x_i)),$$
(5)

where $w(.)$ is the monotonically increasing function that assigns weight to each pairwise violation, and $r_i^u$ denotes the rank of positive label $u$, and $f_u(x_i)$ is the $u\_th$ element of $f(x_i)$. The concept here is that if a lower rank is assigned to the positive label, a greater penalty should be imposed for the violation. However, subsequent research by [59] pointed out that the WARP loss is non-smooth, making it challenging for optimization. Furthermore, it highlighted that the ranking objective falls short of fully optimizing the multi-label objective. Addressing this concern, [59] suggests a new loss function tailored for pairwise ranking based on the LogSumExp function. Their proposed method serves as a smooth substitute for the traditional hinge loss, offering smoothness across all points and facilitating optimization. Other cost functions frequently employed in DL for MLC tasks include the triplet loss function[70], utilized for grouping images into identical label sets, and a resilient logistic loss function [71], which is employed to train CNNs from user-supplied tags.

In 2014, Yoon Kim [72] introduced a text-based model that employs a CNN architecture for text classification and subsequently employs another CNN for sentence-level classification. However, a limitation of this model lies in its inability to overcome the drawback associated with fixed windows in CNNs, thereby hindering its capacity to model long sequence information effectively. Later [73] proposed an XML-CNN model that improved the TextCNN model [72] by incorporating dynamic pooling, refining the loss function with the binary-cross-entropy, and introducing a hidden layer between the output layer and pooling layer. This extra layer is designed to transform high-dimensional labels into a lower-dimensional space, thereby mitigating the computational burden.

In [74], the authors introduced the CNN Hypothesis Pooling (HCP) as an innovative method. It involves utilizing a collection of object segment hypotheses as inputs. Then, each hypothesis is incorporated into a CNN model, and the resulting outputs from different hypotheses are combined through max pooling to generate multi-label predictions. Fig. 4 shows the architecture of HCP. Weiwei et al. [75] proposed deep CNN for multi-label images, incorporating a novel objective function comprising three components: a max-margin objective, a max-correlation objective, and cross-entropy loss. Their proposed framework aims to optimize the utilization of correlation information among labels. This is accomplished by optimizing the score of labels found in the image relative to those that aren't present, utilizing a pre-defined margin. Learning within a semantic space bolsters the correlation between extracted features and their respective labels. Zhu et al. [76] introduced a Spatial Regularization Network (SRN) which leverages attention maps to grasp both semantic and spatial connections among various labels within image datasets. The SRN produces attention maps for each label and captures the inherent relations via trainable convolutions. In a different approach, Kurata et al. [77] proposed a method for initializing neural networks to exploit label cooccurrence information in the context of multi-label text classification. This method applies CNN-based word embeddings to capture label correlations.

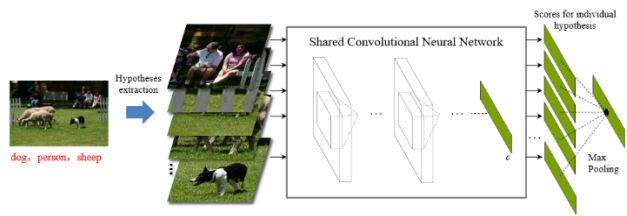

Fig. 4. Hypotheses-CNN-Pooling architecture [74]

In [78], an ensemble of deep CNNs was suggested for multi-label image classification, incorporating well-known architectures such as VGG16 [79] and Resnet-101 [80]. This research explores how different image dimensions impact outcomes and utilizes a range of data augmentation methods alongside a cross entropy loss to train and assess the model. More recently, Park et al. [81] introduced MarsNet, a CNN-based architecture tailored for MLC with the adaptability to inputs of diverse sizes. To handle images of different dimensions, the authors adapted the dilated residual network (DRN) to generate higher-resolution feature maps. Additionally, they introduced horizontal-vertical pooling (HVP) to adeptly amalgamate positional details from these feature maps. The approach further incorporated a multi-label scoring module and a threshold estimation module for MLC, and its effectiveness was validated through a series of diverse experiments. Table I summarizes the most recent CNN/DNN-based approaches for MLC.

MLC can also be executed through the use of a joint label embedding. An example of this is multi-view canonical correlation analysis [82], which is a three-way canonical analysis that aligns the images, labels, and semantics within a shared latent space. Techniques such as WASABI [83] and DEVISE [84] employ a learning-to-rank framework with WARP loss to develop a joint embedding. Metric learning [96] focuses on acquiring a discriminative metric to gauge the similarity between images and labels. Additionally, label encodings can be achieved through methods like matrix completion [97] and bloom filter [98]. While these strategies effectively utilize the semantic redundancy of labels, they frequently fail to capture the dependency of label cooccurrence effectively. Recognizing this limitation, the graph convolution network (GCN) [99] has been found effective in modelling label correlation in the MLC problem. Graph-based deep networks, like graph convolutional neural networks (GCN), offer an effective modelling approach for label dependencies. In this framework, each label is depicted as a node within the graph. Chen et al. [100] propose a directed graph for object labels, employing Graph Convolutional Networks (GCN) for finding correlations between labels. This method translates label representations to interdependent object classifiers, thereby enhancing the overall understanding of relationships between labels. In the same way, semantic-based graph learning [101] incorporates interaction modules and semantic decoupling for associating semantic-based features. The associations are established through GCN constructed from label-correlated data. In a related vein, a subsequent work [102] enhances label awareness by introducing lateral connections between GCN and CNN layers at different depths. This integration ensures improved injection of label information into the backbone CNN. In [103], a deep learning model is proposed to tackle the multi-label patent data challenge, by framing it as a text MLC problem. Their approach involves leveraging GCN to grasp intricate details. This model integrates a dynamic second order attention layer crafted to capture extensive semantic relationships within textual content.

TABLE I
DNN/CNN-based MLC methods with their applications and datasets.

| Ref | Year | Technique | Application | Datasets | Evaluation |
|---|---|---|---|---|---|
| [85] | 2023 | Transfer Learning (VGG16, ResNet) | Movie Genre Prediction | IMDB dataset | Accuracy, AUC, F1-score, Hamming loss |
| [86] | 2023 | CNN based model | Emotion classification | COVID-19 Tweets | F1-score, Accuracy |
| [87] | 2023 | DNN based model | Music classification | OpenMIC Dataset | ROC-AUC, PR-AUC, F1-score |
| [88] | 2023 | Explainable CNN | ECG signals classification | Collection of ECG records | Hamming loss, accuracy, |
| [89] | 2023 | YOLOv5 model | Image Classification | MS-COCO Dataset | Mean average Precision |
| [90] | 2022 | Graph CNN | Image classification | MS-COCO and VOC2007 | Mean average Precision, F1-score |
| [91] | 2022 | YOLOv4 Model | Waste detection | Waste image dataset | Mean average Precision |
| [7] | 2022 | DNN based a | Clinical profile identification | DEFT 2021 dataset | micro-F1 score |
| [92] | 2022 | SHO-CNN | News Classification | RCV1-v2, Reuters-21578, | Hamming loss, F1-score |
| [93] | 2022 | Multi-branch neural network | Image classification | Amazon forest, NusWide, Pascal VOC | F1-score, precision-Recall |
| [94] | 2020 | Encoder-decoder model | Text classification | RCV1-v2, AAPD, Ren-CECps | Hamming loss, Micro-F1 score |
| [94] | 2020 | Seq2Seq-based CNN | Text classification | RCV1-v2, AAPD,Ren-CECps | Hamming loss and micro-F1 score |
| [95] | 2017 | DNN with AutoEncoder | Image and text classification | ESPGame, mirflickr, tmc2007, NUS-WIDE | Micro-F1 and Macro-F1 |

Graph-based techniques like conditional random fields [104], dependency networks [105], and cooccurrence matrices [106] offer solutions for handling label dependency and cooccurrence. Additionally, the label model [107] enhances the label set by incorporating common label combinations. However, these methods typically focus on capturing pairwise label correlations and can become computationally intensive, particularly when dealing with a large number of labels. In contrast, the RNN-based model with low-dimensional recurrent neurons presents a more computationally efficient approach for capturing high-order label correlations. More recent related studies include deep CNN for multi-label image classification [93], CNN-based cross-modal hashing methods [108], Improved sequence generation model via CNN [94], one-dimensional CNN (1D CNN) residual and attention mechanism for multi-label ECG recordings, graphical CNN to grasp label interdependencies using correlations among labels [90].

**Remarks**: Several methods have proposed CNN-based techniques for MLC across diverse data modalities. However, deep CNN is particularly renowned for its effectiveness in multi-label image classification, and it has been applied through two main strategies. The first approach involves training the CNN individually for each label in the image, treating the multi-label problem as a series of single-label tasks [74][109][110]. This method often employs multiple local bounding boxes and instances of learning techniques, resulting in improved performance. However, it tends to overlook potential relationships among labels and struggles to assign labels describing the entire image accurately, as it processes only partial information at a time.

The second strategy adopts a holistic approach by extracting global features from the raw image and employing global loss functions that consider multiple labels simultaneously [111]. This method integrates the entire image into the classification task, enhancing the model's ability to assign labels to describe the overall content. For instance, [69], a deep CNN model utilizing a multi-label loss function has been proposed for top-k ranking. This second-order strategy may compute label correlations, including rankings between relevant and irrelevant labels, resulting in good generalization. Although the second-order strategy harnesses label correlations to a degree, in real-world scenarios, label associations may extend beyond second-order relationships. This can be addressed by a high-order strategy, where MLL considers relations among labels beyond pairwise correlations. This can involve addressing connections among random subsets of labels to capture more complex relationships [55].

## B. LSTMs and Transformers for MLC

A recurrent neural network (RNN) extends a regular feedforward neural network, enabling it to manage variable-length sequential data and undertake time-series prediction. RNNs can be viewed as an extension of the hidden Markov model, integrating a nonlinear transition function, and enabling the modelling of long-term temporal dependencies. LSTM, a variant of RNN, addresses the issue of vanishing gradients by design. It has significantly elevated the field of machine translation, speech recognition, and various other tasks. LSTMs have proven valuable for tasks that were traditionally non-sequential, such as MLC [112][113].

### 1) LSTM based MLC

To our knowledge, the work by Nam et al. [114] marked the first use of RNN to replace the classifier chains for sequence-to-sequence (seq2seq) text classification, thereby effectively capturing label correlations in MLC. Following this, various models have been emerged to address MLC, such as Attentive RNN [115], Orderless RNN [113], and LSTM [116].

LSTM has been applied to solve MLC tasks. Authors in [116] introduced a document classification model leveraging LSTM for multi-label ranking. This model restructures the order of labels within documents to align with a semantic tree. In a similar domain, Yang et al. [117] approach an MLC problem by framing it as the task of sequence modelling. They employ a learning model for sequence synthesis characterized by a decoding architecture to tackle the challenges associated with MLC (Fig. 5). In their sequence generation model, they conceptualize the MLC task as searching for an optimal label sequence $y^*$, aiming to maximize the conditional probability of $p(y|x)$. It is determined in Eq. (6) for a given text sequence $x$, and a subset $y$ containing n labels.

$$p(y|x) = \prod_{i=1}^{n} p(y_i|y_1, y_2, \ldots, y_{i-1}, x) \quad (6)$$

The decoder uses an LSTM to produce labels sequentially, forecasting the next label by leveraging the previously predicted ones. This mechanism allows the model to comprehend label relationships by navigating through label sequence dependencies within the LSTM framework. By adopting a decoder structure in the sequence generation model, it not only discerns correlations among labels but also autonomously picks the most pertinent words while predicting diverse labels.

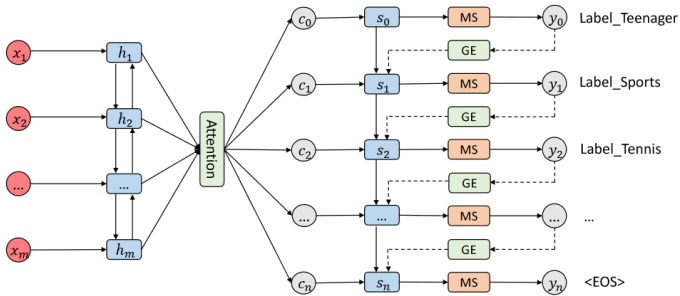

Fig.5. Sequence generation framework. MS refers to the masked SoftMax layer, while GE denotes the global embedding [117].

Yang and colleagues [118] introduced RethinkNet, a DL model designed to overcome the constraints of the CC approach. It tackles the label ordering issue by using a global memory, which retains information on label relationships. This global memory enables all learning models to access the same information, mitigating the label-ordering issue. In another study, researchers [119] proposed a deep RNN for multi-label fault prediction in high-dimensional time series data. Their model incorporates a loss function tailored to account for class imbalance. It comprises two interconnected LSTM networks, namely the encoder and decoder, each geared towards capturing the dynamics of time series data either in the historical or future segments. More recently, Loris Nanni et al. [120] proposed an ensemble method that combines LSTM, GRU, and temporal CNN (TCN) for MLC tasks. Their proposed model undergoes training using various adaptations of Adam optimization, incorporating the concept of Multiple Clustering Centers (IMCC) to enhance the effectiveness of the multi-label classification system. The model employs the binary cross-entropy loss function, expressed by Eq. (7):

$$Loss = \frac{1}{m} \sum_{i=1}^{m} \sum_{j=1}^{l} y_i(j) \cdot \log(h_i(j)) + (1 - y_i(j)) \cdot \log(1 - h_i(j)) \quad (7)$$

where $y_i \in \{0,1\}^l$ and $h_i \in \{0,1\}^l$ represent the actual and predicted label vectors of each sample $((i \in 1, \ldots, m)$, respectively.

Zachary et al. [121] formulated clinical multivariate time series problems into MLC tasks based on LSTM, which marks the first use of LSTM for MLC within the medical domain. The authors employed target replication with LSTM (Fig.6) to effectively classify diagnoses of critical care patients on time series data. The method employing a replication of target produces an outcome $\hat{y}^{(t)}$ at each sequence step. The loss function combines the final loss with the average of losses across all steps in a convex manner (Eq.8)).

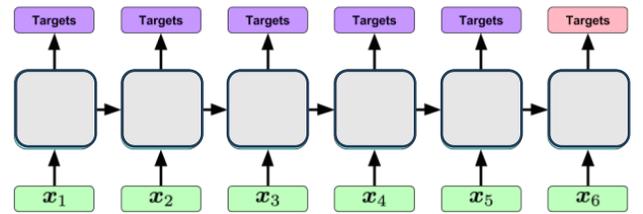

Fig.6. LSTM model with target replication. The main target (red) is used during prediction, while during training, the model back-propagates errors from the intermediate targets (gray) at each sequence step.

$$\alpha \cdot \frac{1}{T} \sum_{t=1}^{T} loss(\hat{y}^{(t)}, y^{(t)}) + (1 - \alpha) \cdot loss(\hat{y}^{(T)}, y^{(T)})$$

(8)

where T represents the total number of sequence steps, and $\alpha \in [0,1]$ is a hyper-parameter that dictates the relative importance of reaching these intermediate targets.

The authors in [122] applied LSTM and Bayesian decision theory for multi-label lncRNA function prediction. They used LSTM for capturing the hierarchical relationships and Bayesian to change the hierarchical multi-label classification problem to the conditional risk minimization problem to obtain final prediction results. Sagar et al. [123] proposed LSTM autoencoder-based multi-label classification for non-intrusive appliance load monitoring. Their proposed method takes electricity consumption as input from the smart meter and reconstructs a time-flipped version of the input using an encoder-decoder paradigm. In the context of multi-label emotion classification, a proposal by [124] presents latent emotion memory (LEM) to acquire latent emotion distribution without relying on external knowledge. LEM comprises latent emotion and memory modules designed to grasp emotion distribution and emotional features, respectively. The combination of these two components is then input into a Bi-directional Gated Recurrent Unit (BiGRU) for the purpose of making predictions.

Other RNN-based studies on MLC include extreme MLC [125], which uses stacked BiGRU for text embedding and integrates attention mechanisms sensitive to clusters to leverage correlations among the large label space. Li et al. [115] propose an end-to-end RNN for weakly-supervised MLC, and [122] designed hierarchical MLC based on LSTM and Bayesian decision theory for LncRNA function prediction.

### 2) AutoEncoder-based MLC

The autoencoders are unsupervised feature representation learning techniques [126] that aim to approximate the input representation by coordinating encoder and decoder layers. These techniques have found extensive application in MLC tasks. Notably, the Canonical Correlated AutoEncoder (C2AE) [95] stands out as the first DL-based label embedding method for MLC. Its structure is shown in Fig. 7. The fundamental concept behind C2AE involves exploring a profound latent space to concurrently incorporate instances and their associated labels. C2AE achieves this by engaging in feature-aware label encoding and leveraging label correlations for accurate predictions. This involves a dual approach: firstly, DCCA (deep canonical correlation analysis) in conjunction with the encoding stage of an autoencoder for feature-conscious label encoding; secondly, introducing a tailored loss function to enhance label-correlation-aware prediction from the decoding outputs. C2AE consists of two key components: DCCA and an autoencoder, both aimed at unveiling three crucial mapping functions: encoding $F_e$, decoding $F_d$, and mapping functions $F_x$. During the training phase, C2AE takes input instances along with the corresponding target labels. C2AE is defined objectively in Equation 9.

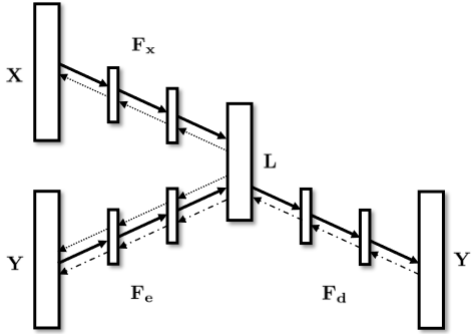

Fig. 7. Architecture of C2AE, which acquires a latent space (L) through nonlinear mappings of $F_x$, $F_e$, and $F_d$. X is the input, and Y is the label data.

$$\Theta = \min_{F_x, F_e, F_d} \Phi(F_x, F_e) + \alpha \Gamma(F_e, F_d) \quad (9)$$

Where $\Phi(F_x, F_e)$ and $\Gamma(F_e, F_d)$ denote the losses in the latent and output spaces respectively with $\alpha$ serving to balance these terms. Drawing upon the principles of CCA, C2AE refines its deep latent space by enhancing the correlation between data instances and their respective labels. Consequently, $\Phi(F_x, F_e)$ can be defined as:

$$\min_{F_x, F_e} \|F_x(X) - F_e(Y)\|_F^2$$

$$s.t \quad F_x(X)F_x(X)^T = F_e(Y)F_e(Y)^T = I \quad (10)$$

where $F_x(X)$ and $F_e(Y)$ represent the transformed feature and label data within the resulting latent space L, respectively.

Later, Bai et al. [127] discovered that the learned deterministic latent space in C2AE lacks smoothness and structure. Minor perturbations within this latent space can result in vastly different decoding outcomes. Despite the proximity of the corresponding feature and label codes, there is no assurance that the decoded targets will exhibit similarity. In order to tackle this issue, [127] proposes an innovative framework, the Multivariate Probit Variational AutoEncoder (MPVAE), designed to acquire latent embedding spaces efficiently and capture label correlations in MLC. MPVAE adeptly learns and aligns two probabilistic embedding spaces—one for labels and another for features. The decoder in the MPVAE framework processes samples from these embedding spaces, effectively modelling the joint distribution of output targets using a multivariate probit model, which is achieved through the learning of a shared covariance matrix. Similar concepts are present in [128], which presents dual-phase label embedding (TSLE) through a neural factorization machine for MLC, as depicted in Fig.11. Within this framework, the encoder is a Twin Encoding Network (TEN) composed of a singular feature network and a singular label network. The decoder's objective is to reconstruct the label based on the feature embedding. Both the feature network and the label network employ a factorization layer, enabling the computation of pairwise correlations between features and labels.

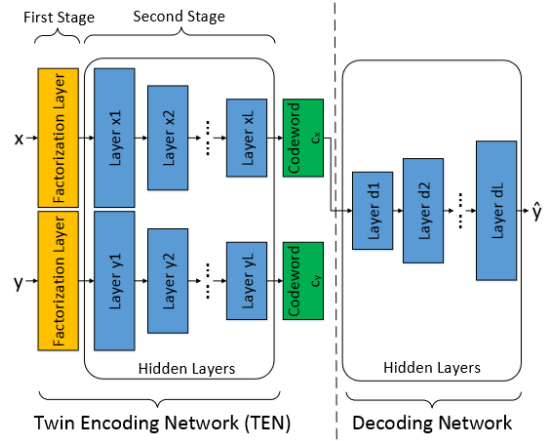

Fig. 11. Illustrates the structure of the TSLE paradigm [128].

Recently, Bai et al. [129] proposed an extension of their previous MPVAE [127] model by proposing contrastive learning boosted Gaussian mixture variational autoencoder (C-GMVAE). The loss function employed in their model is a combination of several components, including the KL loss of feature and label embeddings, the VAE reconstruction loss, a supervised contrastive loss of feature and label embeddings, and, ultimately, a cross-entropy loss for classification. In [130], authors propose a autoencoder with dual-encoding layer aimed at exchanging knowledge through the second encoding weight matrix. The autoencoder model is designed to jointly optimize representation learning and multi-label learning to enhance MLC performance. Nevertheless, existing methods based on autoencoders typically rely on a single autoencoder model, presenting challenges in the learning of multi-label feature representations and lacking the capability to assess similarities between data spaces.

To tackle the single model limitation of [130], Zhu et al. [131] propose a new approach called the RLDA (representation learning method with dual autoencoder). This method effectively captures diverse characteristics and abstract features from data through the sequential integration of two distinct autoencoder types. First, the algorithm utilizes reconstruction-independent component analysis (RICA) within a sparse autoencoder framework, training on patches from both the training and test datasets to learn global features robustly. Following this, the output from RICA is utilized, and then a stacked autoencoder with manifold regularization is employed to refine the quality of multi-label feature representations. Ultimately, the sequential combination of these two autoencoder types generates innovative feature representations for multi-label classification.

### 3) Transformer based MLC

Originally introduced for capturing long-term dependencies in sequence learning challenges, transformers [132] have found extensive application in various natural language processing tasks. More recently, models based on transformers have been developed for numerous vision-related tasks, demonstrating significant promise in the field. The application of transformers in addressing MLC arises from the necessity to dynamically extract local discriminative features tailored to different labels. This adaptive feature extraction is a highly desirable property, particularly in scenarios involving multiple objects within a single image.

Ramil and Pavel [133] were the first to apply the BERT model to the multi-label problem and investigate its efficacy in hierarchical text categorization challenges. They introduced a BERT-based model aimed at generating sequences for multi-label text classification. Gong et al. [134] later introduced an HG-transformer, a deep learning architecture that first converts input text into a graph structure. This model then employs a multi-layer transformer with a multi-attention mechanism at the word, sentence, and graph levels to capture text features comprehensively. Utilizing hierarchical label relationships, the model generates label representations and incorporates a weighted loss function tailored to semantic label distances. While the effectiveness of the transformer-based MLC model surpasses that of CNN and RNN structures, it is noteworthy that transformer models often involve a substantial number of parameters and a complex network structure, leading to practical limitations. In the pursuit of enhancing the applicability of transformers in MLC, [135] presented a hybrid model named tALBERT, which merges LDA and ALBERT to derive diverse multi-level document representations. Extensive experiments conducted using three datasets substantiate the higher performance of their hybrid method compared to the current state-of-the-art methods in the realm of multi-label text classification.

The framework Query2Label, introduced in the study [136], presents a novel approach to MLC by employing a transformer decoder. To our knowledge, this marks the first application of such a framework to address MLC challenges. Query2Label operates in two stages, utilizing Transformer decoders to extract features. The multi-head attention employed in this process focuses on distinct facets or perspectives of an object category. Furthermore, the framework autonomously learns label embeddings from the provided data. To better handle the imbalance problem, the framework adopts a simplified asymmetric focal loss for computing the loss of each training sample, as shown in Eq. (12).

$$L = \frac{1}{K}\sum_{k=1}^{K}\begin{cases}(1-p_k)^r + \log(p_k), y_k = 1 \\ (p_k)^r - \log(1-p_k), y_k = 0\end{cases} \quad (12)$$

where $y_k$ represents a binary label that indicates whether the image $x$ is associated with label $k$, the overall loss is calculated by averaging this specific loss across all samples within the training dataset. $r$ denotes the focal parameter, and stochastic gradient descent is employed for optimization. The default values for $r+$ and $r-$ are set to 0 and 1, respectively.

Ridnik et al. [137] propose an ML-Decoder model, which can provide a unified solution for multi-label classification. The ML-Decoder can be seamlessly applied by excluding any pre-trained, fully connected layers, demonstrating a consistently improved balance between speed and accuracy in tests conducted on the MS-COCO MLC task. [138] A triplet transformer architecture designed for multi-label document classification was presented, as shown in Fig. 9.

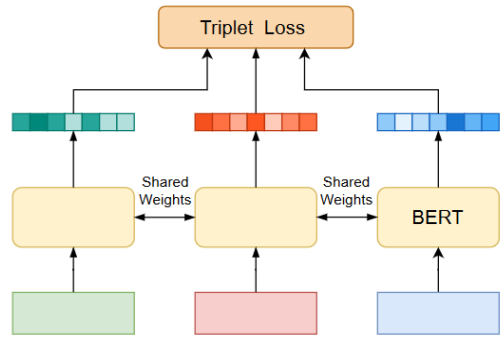

Fig. 9. Triplet transformer network for MLC [138].

This model is adept at embedding both labels and documents into a unified vector space. The architecture comprises three

BERT Networks with shared weights, facilitating document classification by pinpointing the nearest and, consequently, the most akin labels. More recently, [139] proposed a graph attention transformer network (GATN) for the MLC problem. This network is specifically designed to efficiently discover intricate relationships among labels. GATN employs a two-step process to enhance the expressive capacity of label relations. Initially, cosine similarity is applied using label word embeddings to create an initial correlation matrix, capturing extensive semantic information. Subsequently, to adapt this adjacency matrix, a graph attention transformer layer is crafted to the current domain. They adopted node embeddings to construct their final correlation matrix, as shown in Eq. (12). The generation of the correlation matrix is performed as follows. For an MLC having n categories, let $Z_i$ represent the embedding vector of the i-th label of length d and $R_{i,j}$ denote the correlation value between the i-th label and the j-th label. $R_{i,j}$ is calculated based on the cosine similarity of the embedding vectors (Eq. 13).

$$R_{i,j} = \frac{\sum_{k=1}^{d} Z_{i,k} \times Z_{j,k}}{\sqrt{\sum_{k=1}^{d} (Z_{i,k})^2} \times \sqrt{\sum_{k=1}^{d} (Z_{j,k})^2}} \quad (13)$$

The resulting matrix R is symmetric, where $R_{i,j}$ is always equal to $R_{j,i}$, indicating a consistent relationship between any two categories. By using a threshold $\tau$ and weight parameter $p$, the ultimate correlation matrix A can be derived.

$$R'_{i,j} = \begin{cases} 0, if\ R_{i,j} < \tau \\ 1, if\ R_{i,j} \geq \tau \end{cases}$$

$$A_{i,j} = \begin{cases} p/\sum_{j=1}^{n} R'_{i,j}, if\ i \neq j \\ 1 - p,\ if\ i = j \end{cases} \quad (14)$$

In a different study, Chen et al. [148] developed multi-label image recognition based on spatial and semantic transformers (SST). SST functions as a modular plug-and-play system capable of simultaneously extracting semantic and spatial correlations within multi-label images. It consists of two independent transformers with distinct objectives: the spatial transformer focuses on modelling correlations among features from various spatial positions, whereas the semantic transformer is employed to apprehend the coexistence of labels without the need for manually defined rules. More recently, large language models pre-trained on large datasets using transformer architectures have been suggested for multi-label text classification, such as LP-MTC [42], prompt tuning [149], and SciBERT [150]. Table II summarizes state-of-the-art transformers and LSTMs adopted to address MLC challenges.

**Remarks**: Transformers and autoencoders have become some of the most successful DL approaches actively adopted for MLC in recent years. These techniques have been found in various real-world scenarios, including multi-label emotion classification [4], MLC in video for underwater ship inspection [151], MLC in textual data [135], and multi-label disease classification [152]. MLC models built on transformer structures often outperform those based on RNN and LSTM. However, it's worth noting that transformer models typically entail a substantial number of parameters and a complex network structure, introducing certain limitations in practical applications. Moreover, addressing label correlations in MLC is crucial for certain objectives, presenting a challenge due to the inherent structure of the label space within the data. In future studies, exploring effective methods for capturing label correlations and other related challenges will be the main research focus of autoencoders and transformers for MLC.

*C. Hybrid DL for MLC Problem*

One common approach to extending CNNs to MLC involves transforming the problem into multiple single-label classification tasks, which use either ranking loss [69] or cross-entropy loss [153]. However, these methods fall short of capturing the dependencies between multiple labels when treating them independently. Several studies have demonstrated the significant label cooccurrence dependencies in MLC problems. To capture label dependency, existing studies have employed a range of techniques including methods based on nearest neighbour [154][155], ranking-based methods [156][157], structured inference models [158][159], and graphical models [106][160][107]. A prevalent strategy involves representing dependencies and cooccurrence relationships through pairwise compatibility probabilities or cooccurrence probabilities. Subsequently, Markov random fields [105] are often utilized to deduce the final joint probability of labels. However, in scenarios involving numerous labels, the parameters linked with these pairwise probabilities can grow disproportionately large, potentially containing redundancies, especially if labels exhibit considerable overlap in meaning. Moreover, many of these approaches either lack the capability to capture higher-order correlations [106] or necessitate increased computational complexity to accommodate more nuanced label associations [107]. In the last few years, there has been a shift towards RNN, particularly LSTM [161] following CNN, as an effective method to exploit high-order label correlations. LSTMs exhibit the ability to capture complex inter-label relationships while still maintaining manageable computational complexity.

TABLE II
Review of state-of-the-art LSTM/transformer-based approaches to MLC applications.

| Ref | Year | Technique | Application | Datasets | Evaluation |
|---|---|---|---|---|---|
| [140] | 2023 | BERT based models | Toxicity content identification | Online Corpus | Precision, recall, and F1-score |
| [3] | 2023 | Transformers | Document classification | AAPD dataset, Reuters-21578 | F1-score, Precision, recall |
| [141] | 2023 | Transformers | Chest X-ray diagnosis | PadChest dataset | AUC, and mean AUC |
| [142] | 2023 | Transformer network | Retinal Disease Classification | MuReD, ARIA, STARE, etc. | Precision, recall, F1, AUC |
| [143] | 2022 | LSTM with GloVe method | Cardiovascular disease classification | Cardiovascular text dataset | Accuracy |
| [144] | 2022 | Transformer model | Image Classification | MS-COCO, Pascal-VOC, NUSWIDE | Precision, recall, F1-score |
| [122] | 2022 | LSTM Network | LncRNA function prediction | GOA-lncRNA dataset | Micro F1, Macro F1 |
| [145] | 2022 | Transformer-CNN | Text classification | RCV1 and AAPD | Micro and macro F1 |
| [146] | 2022 | encoder–decoder | Text classification | AAPD and RCV1-V2 | Hamming loss, Micro-F1 |
| [147] | 2021 | Deep graph LSTM | Legal Text classification | Legal case of Indian judiciary | Accuracy |

The concept of exploiting RNN models to address label dependencies in MLC was initially introduced in [162] and [163], where the fusion of CNN with RNN architecture was proposed. Because classifier chains (CC) are considered a memory mechanism that stores the label predictions of the earlier classifiers, CNN-RNN-based algorithms can extend CC by replacing the mechanism with a more sophisticated memory-based model. Wang et al.[162] put forward a unified CNN-RNN framework that learns a joint image-label embedding to characterize semantic label dependencies. The CNN-RNN structure comprises a CNN feature mapping layer (encoder) for extracting semantic representations from images and an RNN inference layer (decoder) that utilizes the encoding to generate a label sequence, modelling image/label relationships and dependencies. RNNs adopt a frequent-first ordering approach for sequential outputs, and multiple label outputs are generated at the prediction layer through the nearest neighbour search. In [163], it was demonstrated that the sequence of labels during training significantly influenced annotation performance, with the rare-to-frequent order producing the most favourable outcomes, a finding corroborated in subsequent studies like [162][164]. Jin et al. [107] utilized CNN to represent images, feeding them into an RNN for predictions. They experimented with rare-first, dictionary-order, frequent-first, and random label ordering, comparing the outcomes of each method. Liu et al.[164] employ a comparable framework, wherein they explicitly assign the tasks of label prediction and label correlation to the CNN and RNN models, respectively. Rather than employing a densely connected layer linking the RNN and CNN models, the researchers feed the RNN with output features extracted from the CNN model, thereby supervising both models throughout training. They adopt a rare-first ordering in their model to accord greater significance to rare labels. The authors explore various visual representations to input into the RNN. In [162], the exploration of the image-text relationship involves mapping images and labels into a common low-dimensional space, while [164] employing predicted class probabilities and [163] investigating various internal layers of the CNN in its experiments. Other works based on CNN-RNN architecture include an ensemble of CNN-RNN for text categorization [165], CNN-RNN for satellite images of the Amazon rainforest [166], hybrid CNN, and bidirectional LSTM network for multi-label aerial image classification [167] and CNN-ConvLSTM for pedestrian attribute recognition [168], as shown in Fig. 10.

Despite the encouraging performance exhibited by the CNN-RNN architecture, its reliance on a predetermined label order for learning poses a significant challenge. Since RNN-based models produce sequential outputs, a pre-defined label order is required during training for the MLC task. For example, Wang et al.[162] determine the label order based on label frequencies observed in the training data. However, employing such predetermined label orders may not accurately capture natural label dependencies, introducing a rigid constraint on the RNN model. Introducing a frequent-to-rare label order skews the model towards prioritizing frequent labels, necessitating numerous accurate predictions before addressing the rare label(s). Conversely, a rare-to-frequent label order compels the model to prioritize learning rare labels, a challenging task given the scarcity of training examples. Overall, a frequency-based pre-defined label order typically overlooks genuine label dependencies. This is because each label in a multi-labelled image is intricately linked to many other labels within the broader context, even though a label may display a stronger association with only a select few. Moreover, defining such orders introduces a bias towards dataset-specific statistics, undermining the generalizability of the model. The lack of robustness in learning optimal label orders, as verified in [162], is exacerbated by the difficulty in predicting labels for smaller-sized objects when visual attention information is underutilized. Consequently, addressing how to introduce flexibility in learning optimal label orders while concurrently exploiting associated visual information becomes an important focus of research.

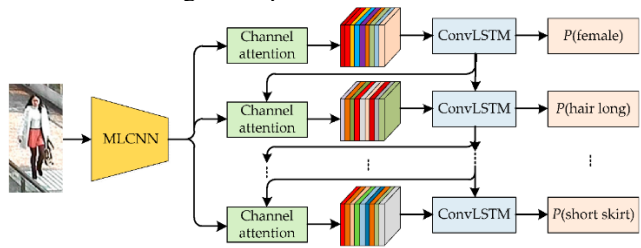

Fig.10. CNN- ConvLSTM for pedestrian attribute recognition [168].

In resolving these constraints related to the order of labels, some studies [112][113] have suggested methods that eliminate the necessity of feeding ground truth labels to the RNN in a predefined sequence. Chen et al. [112] introduce an RNN for MLC that doesn't rely on order, incorporating visual attention and LSTM models. They utilize binary cross-entropy loss at each time step to predict labels without considering their order. The simultaneous learning of attention and LSTM models enables the identification of regions of interest associated with each label, automatically capturing label order without pre-defined constraints. The integration of attention with LSTM results in slight enhancements over CNN-RNN models. Additional work presented in [113] proposes an order-less recurrent model for MLC, as depicted in Fig. 11. This approach explores how predicted label sequences interact with the dynamic ordering of ground truth labels, enabling the expedited training of LSTM models for greater optimization. Notably, this method avoids duplicate generation, contrasting with Chen et al.'s [112] explicit duplication removal module. Subsequent studies, including [169] and [115], have also embraced order-free strategies in MLC. In [169], the authors propose an end-to-end trainable framework for multi-label image recognition. This framework comprises CNN for extracting intricate feature representations, alongside an attention-aware module utilizing LSTM. This module iteratively identifies regions relevant to classes and predicts label scores for these discerned regions. The end-to-end training of the framework relies solely on image-level labels facilitated by reinforcement learning techniques. The process is initiated by inputting the image into the CNN, yielding feature maps. Subsequently, an LSTM module processes those features, incorporating the hidden state from the prior iteration to predict scores for each region. These scores play a pivotal role in determining the location for the subsequent iteration. To arrive at the ultimate label distribution, the predicted scores undergo consolidation through category-wise max-pooling. However, these methods tend to internally select a specific label order at the initial time step and then continue to follow the same sequence in subsequent time steps. In essence, these strategies enable the RNN to implicitly favour one among numerous sequences, thereby introducing inherent bias.

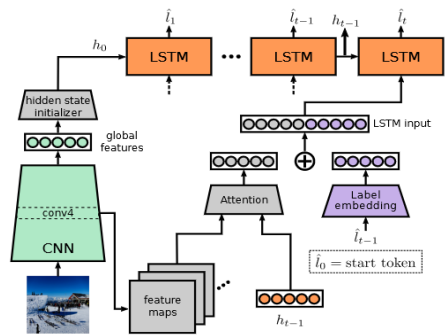

Fig.11. Illustration of CNN-LSTM architecture where the orderless loss function substitutes the loss function [113].

Later, Ayushi et al. [175] proposed multi-order RNN to address the limitation of these order-free approaches (Fig.12). Their approach provides RNNs with the flexibility to explore and grasp various pertinent inter-label correlations through multiple label orders, rather than being constrained to predetermined and fixed one. The architecture of multi-order RNN comprises a deep CNN fine-tuned with ground-truth data from a specified dataset and an LSTM model that leverages the soft confidence vector obtained from the CNN as its starting point. At each time step within every sample, a cross-entropy loss is calculated, considering all true labels except the one from the preceding time step as potential predictions. The ultimate predictions are derived by max-pooling individual label scores across all time steps. Demonstrating superior performance over CNN-RNN methods, the multi-order RNN offers an intuitive approach to adapting sequence prediction frameworks for image annotation tasks.

More recently, Wang and his colleagues [176] introduced a novel approach for multi-label image classification. Their method integrates cross-modal fusion with an attention mechanism, leveraging both graph convolution networks and attention mechanisms to handle both local and global label correlations seamlessly. This approach comprises three main components: a module for feature extraction equipped with an attention mechanism, a module for learning label co-occurrence encodings, and a cross-modal fusion module employing Multi-modal Factorized Bilinear pooling. Through efficient fusion of image features and label co-occurrence embeddings, their method demonstrates promising results. Their method was tested on COCO and VOC2007 datasets and showed better classification results compared to other similar approaches. A different study [177] introduced two hybrid DL models for classifying DNA sequences. The first model combines SCAE (stacked convolutional autoencoder) with MLELM ( Multi-label Extreme Learning Machine), while the second model incorporates VCAE (variational convolutional auto-encoder) into the MLELM framework. These models adeptly produce accurate feature maps that capture both individual and inter-label interactions within DNA sequences, encapsulating spatial and temporal characteristics. The extracted features are fed into MLELM networks, yielding soft classification scores and hard labels. Notably, the VCAE-MLELM model consistently outperformed the SCAE-MLELM model, while the latter demonstrated superior performance in soft categorization, surpassing existing methods, such as CNN-BiLSTM and DeepMicrobe [178]. The convolutional autoencoder aids in extracting spatial organization, enhancing computational efficiency by identifying latent associations. The extracted features are input into two MLELMs, where the first model produces probabilistic labels, and the following model establishes associations between deterministic and probabilistic labels. Table III summarizes some of the state-of-the-art Hybrid DL techniques in MLC, along with their domain of application, datasets, and evaluation metrics.

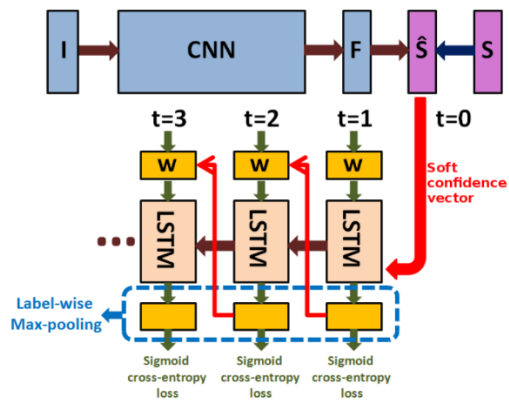

Fig. 12. Multi-order RNN for Image Annotation

TABLE III
Review of state-of-the-art hybrid-based approaches to MLC applications.

| Ref | Year | Technique | Application | Datasets | Evaluation |
|---|---|---|---|---|---|
| [170] | 2023 | Transformer + CNN | Medical imaging | ChestX-ray11, NIH ChestX-ray14 | Mean avg. Precision, recall, F1 |
| [171] | 2023 | CNN-BiLSTM | Short texts Classification | Online large corpuses | Micro-f1, Macro F1 |
| [146] | 2023 | Hybrid transformer | Text Classification | AAPD and RCV1-V2 | Hamming loss, F1 |
| [172] | 2022 | AutoEncoder+CNN | Movie and image | IMDb Movie, MS-COCO | precision, Recall, Hamming loss |
| [120] | 2022 | LSTM+GRU+TCN | Music, image, drug | Cal500, Image, Scene, Yeast, etc. | Hamming loss, Ranking loss, |
| [173] | 2021 | Graph CNN + CNN | Breast cancer | Mammograms 2018 | Accuracy, TPR |
| [174] | 2019 | Ensemble of CNN | Image classification | NUS-WIDE, MS-COCO, PASCAL | Mean average Precision, F1 score |
| [162] | 2017 | CNN +RNN | Image classification | NUS-WIDE, MS-COCO, VOC PASCAL 2007 | per-class and overall (recall, Precision, etc.) |

## V. MLC CHALLENGES AND DATASETS

The sustained appeal of MLC can be attributed to the widespread prevalence of multi-label data, which is pervasive across various domains, such as biology, environment, healthcare, commerce, recommender systems, social media, retail, sentiment analysis, energy, transportation, and robotics. Furthermore, the internet consistently generates quintillions of bytes of streaming data daily, posing significant challenges for MLC tasks. In real-world applications, MLC continues to pose challenges owing to the intricate nature of label presence. For instance, in certain cases, there exist scenarios characterized by a vast number of labels, where these labels are only partially or weakly provided, and their emergence could be continuous or entirely unforeseen. This section examines some of the challenges in MLC and the datasets associated with it.

- **Label Dependencies:** The presence of multiple data labels can suggest associations between different entities. For instance, in the task of identifying objects within images, dogs and cats may frequently coexist, while cats and sharks typically do not share the same space. Thus, modelling and learning correlations among categories have consistently remained a fundamental focus in MLC [190]. However, effectively leveraging label dependencies continues to be a persistent challenge in MLC. Regarding the modelling of these dependencies, learning approaches can be classified as first-order (addressing each distinct label independently), second-order (modelling pairs of labels), and higher-order (simultaneously addressing more than two labels). The powerful learning capabilities of DL methods are commonly harnessed to tackle second-order label dependencies in various ways, including through graph CNN [90][191], autoencoder-based [127][129], transformers [3][192], and hybrid DL models [170] [171] [162]. However, the challenge of higher-order label dependence in MLC continues to be a central focus for researchers, including both practical and theoretical considerations, persisting into the present day.
- **Extreme MLC:** Another important challenge in MLC is the presence of a very large number of labels, also known as extreme MLC. It is an active area of research wherein the number of labels can be exceptionally high, reaching into the millions in certain cases. Traditional classifiers like one-vs-all, SVM, and neural networks face two primary impediments when applied directly in the context of extreme MLC [193]. Initially, the considerable volume of labels presents a major bottleneck, as implementing a classic model for all labels is impractical due to computational limitations. Additionally, the existence of labels with minimal samples further complicates the learning process for these specific labels. Several efforts have been made, such as Ranking-based Auto-Encoder (Rank-AE)[194], DeepXML framework [195], AttentionXML [24], and two-stage XMTC framework (XRR)[26] to address the challenges posed by extreme MLC.
- **Weakly-Supervised MLC:** Weakly supervised learning focuses on the more demanding aspect of MLC, wherein certain labels in the training set are missing. Given the extensive data volumes and diverse domains involved in such tasks, fully supervised learning demands manually annotated datasets, incurring significant costs and time. The weakly supervised MLC task, involving the training of MLC models with partially observed labels per sample, is gaining importance due to its potential for substantial savings in annotation costs [29]. In addressing MLC with missing or partial labels, several notable approaches have been suggested, including Graph Neural Networks (GNNs) [33], deep generative models[30], and hierarchical MLC [196]. Moreover, learning paradigms, such as zero-shot learning [197], few-shot learning [198], and self-supervised learning [199] are emerging research directions for partial or weakly supervised MLC.

TABLE IV
Multi-label datasets and their description

| Datasets | Domain | #Labels | #Features | #Instances | Cardinality | Density | Diversity |
|---|---|---|---|---|---|---|---|
| Bibtex [179] | Text | 159 | 1836 | 7,395 | 2.402 | 0.015 | 0.386 |
| EUR-Lex [180] | Text | 3993 | 5000 | 19,348 | 1.292 | 0.003 | 0.083 |
| RCV1[181] | Text | 103 | 100000 | - | - | - | - |
| Corel5k [182] | Images | 374 | 499 | 5000 | 3.522 | 0.009 | 0.635 |
| NUS-WIDE [183] | Images | 81 | 500 | 269,648 | 1.869 | 0.023 | - |
| AmazonCat-13K | Text | 13330 | - | 1.5 M | - | - | - |
| CNIPA-data [61] | Text (patent) | 618 | 17/242 | 212,095 | 1.330 | 0.002 | - |
| USPTO-data [61] | Text (patent) | 632 | 8/111 | 353,701 | 2.174 | 0.003 | - |
| Birds [184] | audio | 19 | 260 | 645 | 1.014 | 0.053 | - |
| CAL500 [185] | Music | 174 | 68 | 502 | 26.044 | 0.150 | 1.000 |
| Emotions [186] | Music | 6 | 72 | 593 | 1.869 | 0.311 | 0.422 |
| CMU-Movie [61] | Movie | 372 | 3/312 | 42,204 | 3.691 | 0.010 | - |
| Liu [187] | Drugs | 1385 | 2892 | 832 | - | - | - |
| Yeast [111] | Biology | 14 | 103 | 2,417 | .237 | 0.303 | 0.082 |
| Genbase [188] | Biology | 27 | 1186 | 662 | 1.252 | 0.046 | 0.048 |
| Water quality [189] | Chemistry | 14 | 16 | 1,060 | 5.073 | 0.362 | - |
| GoEmotions [61] | Emotion | 28 | 13 | 211,225 | 1.181 | 0.042 | - |
| Toxic [61] | Comments | 6 | 70 | 159,571 | 0.220 | 0.037 | - |

- **Imbalanced MLC:** Imbalanced learning is a widely recognized and intrinsic characteristic observed in multi-label datasets, influencing the learning dynamics of various MLC algorithms. The issue of imbalance in MLD can be analyzed from various factors [20]: intra-label imbalance, inter-label imbalance, and imbalance among label sets. These factors may also co-occur, further intensifying the complexity of the MLC task. Although traditional independent approaches are commonly used for addressing imbalanced MLDs [200], DL model adaptation methods are still underdeveloped [201][202].
- **High Data Dimensionality:** Much like numerous learning tasks, MLC faces the challenge of dimensionality. The rapid expansion of data scaling in multi-label datasets often results in high-dimensional features [203], contributing to prolonged processing times and diminished classifier performance. The occurrence of this problem stems from an abundance of redundant, noisy, and irrelevant features, giving rise to overfitting problems. To mitigate these challenges, it becomes imperative to reduce feature dimensionality with two primary approaches: feature extraction and feature selection. The former involves mapping high-dimensional features into a lower-dimensional space [204], while the latter entails choosing a smaller subset of features to replace the entire original set [205]. Feature extraction produces new features that lack physical meaning, whereas feature selection preserves physical meaning and enhances explanatory power. Table IV provides a summary of some multi-label datasets spanning diverse domains. The table includes information such as the number of instances, features, labels, cardinality, density, and diversity. The names listed in Table IV may not necessarily match those mentioned in the original papers;

TABLE V
Comparison of deep learning techniques for solving multi-label classification

| Ref | Neural Network Type used | Label Correlation Strategy used | Data Modality | Advantages | Limitations |
|---|---|---|---|---|---|
| [3] | Transformers | A dependency regularization approach is added to the loss function | Textual/ Document | Effective use of transformers to learn pairwise label co-occurrence and dependencies | Limited to handle effectively higher order label correlation. |
| [90] | Graph CNN | Semantic label embedding | Images | The method can resolve the scalability and label cooccurrence issues in MLC | Limited to handling interdependent relationships among the label |
| [56] | Multi-layer Feedforward DNN | Employing a new error function with pairwise ranking loss | Genomics and Textual | Excel in MLC tasks by accounting for label correlations, thereby improving performance. | Computational complexity with label size. Does not perform well on textual data |
| [95] | DNN with AutoEncoder | Canonical Correlated AutoEncoder (C2AE) | Texts and Images | Allows better exploitation of cross-label dependency during prediction processes | Different objectives are mixed and difficult to understand |
| [163] | CNN-RNN | Based on pre-defined label ordering using frequent-to-rare or rare-to-frequent | Image Datasets | Captures the characteristics of MLC and performs better for sequence generation problem | Pre-set label order fails to capture genuine label interdependencies. |
| [207] | SHO-LSTM | Word embedding | Textual | good convergence capability and optimal solution | Limited to handle label dependency |
| [112] | CNN-RNN | Order-free RNN with Visual Attention | Image Datasets | Doesn't require pre-defined label order for label correlation and prediction | Label label-ordering approach doesn't work for some problems |
| [208] | Transformers | Linguistic and Semantic Cross-Attention | Multi-modal data | Handles high-level semantics and linguistic embeddings | Does not consider label cooccurrences |
| [175] | CNN-RNN | Multi-order RNN | | Multi-order RNN | Computational complexity |
| [120] | Ensemble Model | Ensemble of GRU, LSTM, and TCN | Music, Biological | Better performance and stability in prediction | Computational complexity |
| [58] | Deep NN | Cross entropy loss function | Textual/ Document | Addresses limitation of ranking loss minimization | Doesn't contain a deep-layer network for large-scale image data |
| [77] | Deep NN | Word embeddings based on CNN with binary cross-entropy | Textual datasets | Less computational overhead during training and evaluation | weight initialization requires a sophisticated approach |
| [174] | Deep CNN | Sigmoid cross-entropy loss | Images Dataset | Ability to handle various size inputs with a strong baseline | Doesn't consider the correlation between labels |
| [116] | LSTM2 and rankLSTM | Word embedding based on a hybrid loss function | MEDLINE Dataset | employs a semantic tree for ranking labels and tackles error propagation when dealing with a variable number of labels | Not efficient for full-text document classification. |
| [118] | RethinkNet | Adopt a global memory approach | Emotions, medical, | Tackles both label correlation and class imbalance | Running time complexity |
| [162] | CNN-RNN | Joint image-label embedding for dependency | Imaging dataset | Hybrid use of CNN and RNN makes the task tractable | Label order based on frequency does not accurately depict the genuine correlation among labels. |
| [209] | Hyperspectral CNN | paired network structure and utilizes a hybrid mechanism | Textual data | Reduced imbalanced problem in MLC | Does not capture the relationship between labels |
| [135] | tALBERT (LDA+ ALBERT) | Deep semantic information | Documents | utilized for extracting the depth features from documents | The model is only effective on fewer data instances |

instead, they are the ones commonly used in the literature. To describe the characteristics of a multi-label dataset, three metrics can be employed for measurement, namely, label cardinality, label density, and label diversity [206]. Given a multi-label dataset $M = \{(X_i, Y_i) | 1 \leq i \leq N\}$, the three-attribute metrics can be defined as follows, where m = |M| represents the dataset size, $|Y_i|$ represents the total number of labels for $i^{th}$ instance, and |L| the label count in M.

- **Label cardinality (Card)**: It represents the average number of labels per sample.

$$\text{Card}(M) = \frac{1}{m} \sum_{i=1}^{m} |Y_i| \quad (15)$$

- **Label density (Dense)**: It denotes the average number of labels per sample in dataset M, divided by the total number of labels.

$$\text{Dens}(M) = \frac{1}{m} \sum_{i=1}^{m} \frac{|Y_i|}{|L|} \quad (16)$$

- **Label Diversity (Diver):** It indicates the count of distinct label combinations present in the sample set.

$$\text{Diver}(M) = |\{Y_\times \ \exists \times : (X, Y_\times) \in M\}| \quad (17)$$

The challenge in the MLC problem arises from the high dimensionality of features and labels. Several suggestions have been made regarding dimensionality reduction [205] and feature selection [210] in MLC. A recent contribution by Zan et al. [211] introduces a new method known as global and local feature selection (GLFS), demonstrating its superiority over current advanced techniques for selecting features across multiple labels. However, real-world MLC scenarios present additional complexities in the feature space, such as the potential disappearance or augmentation of certain features and alterations in distribution. Effectively addressing these challenges concurrently in both label and feature spaces represents a more formidable task and constitutes a prospective avenue for future research in addressing the intricate nature of the MLC problem.

## VI. COMPARATIVE ANALYSIS

In this section, we conduct a comparative examination of diverse DL methods suggested in existing literature to tackle MLL problem. The comparison parameters encompass network architecture, approaches to label correlation, evaluation metrics, key findings, and limitations of each proposed method. Table V depicts detailed information on the various methods proposed so far for solving the MLC using DL approaches.

## VII. CONCLUSIONS

Due to its robust learning ability, DL has achieved higher performance in various practical multi-label learning applications, including multi-label image and text classification. In solving multi-label learning problems, the main challenge lies in effectively leveraging DL to capture label dependencies more adeptly. This paper presents an extensive examination of DL methods for MLL problems, with a primary emphasis on DL for multi-label classification (MLC) involving label correlations. We have compiled and scrutinized numerous articles pertaining to DL techniques for MLC published from 2006 to 2023. The survey details recent methods related to various DL approaches, including DNN, CNN, LSTM, autoencoders, transformers, and hybrid models to tackle MLC challenges. The study provides an overview of the representative works cited, delving into the latest DL techniques applied in MLC and scrutinizing their limitations. It also covers a concise depiction of the existing challenges in MLC and a brief description of the publicly available multi-label datasets. Furthermore, we conducted a comparative analysis of DL approaches for MLC, highlighting the strengths and weaknesses of existing methods and offering insights into promising avenues for future research.

Overall, despite the increasing demand for MLC across diverse domains, the research on developing an efficient and comprehensive DL framework for MLC, along with an effective model to address the associated challenges, such as label correlations remains still a challenge. Consequently, there is a need for further exploration to identify more effective solutions in the future.